# Low-latency vision transformers via large-scale multi-head attention

Ronit D. Gross[a,1], Tal Halevi[a,1], Ella Koresh[a], Yarden Tzach[a], and Ido Kanter[a,b,*]

[a]Department of Physics, Bar-Ilan University, Ramat-Gan, 52900, Israel.
[b] Gonda Interdisciplinary Brain Research Center, Bar-Ilan University, Ramat-Gan, 52900, Israel.
*Corresponding author at: Department of Physics, Bar-Ilan University, Ramat-Gan, 52900, Israel.
E-mail address: ido.kanter@biu.ac.il (I. Kanter).
[1]These authors equally contributed to this work

## Abstract

The emergence of spontaneous symmetry breaking among a few heads of multi-head attention (MHA) across transformer blocks in classification tasks was recently demonstrated through the quantification of single-nodal performance (SNP). This finding indicates that each head focuses its attention on a subset of labels through cooperation among its SNPs. This underlying learning mechanism is generalized to large-scale MHA (LS-MHA) using a single matrix value representing single-head performance (SHP), analogous to single-filter performance in convolutional neural networks (CNNs). The results indicate that each SHP matrix comprises multiple unit clusters such that each label being explicitly recognized by a few heads with negligible noise. This leads to an increased signal-to-noise ratio (SNR) along the transformer blocks, thereby improving classification accuracy. These features give rise to several distinct vision transformer (ViT) architectures that achieve the same accuracy but differ in their LS-MHA structures. As a result, their soft committee yields superior accuracy, an outcome not typically observed in CNNs which rely on hundreds of filters. In addition, a significant reduction in latency is achieved without affecting the accuracy by replacing the initial transformer blocks with convolutional layers. This substitution accelerates early-stage learning, which is then improved by subsequent transformer layers. The extension of this learning mechanism to natural language processing tasks, based on quantitative differences between CNNs and ViT architectures, has the potential to yield new insights in deep learning. The findings are demonstrated using compact convolutional transformer architectures trained on the CIFAR-100 dataset.

## 1. Introduction

Two primary classes of architectures are available for solving complex image-classification tasks. The first class, convolutional neural networks (CNNs) [1-3], was introduced several decades ago and is based on progressively expanding receptive fields along the convolutional layers (CLs). The second class, introduced more recently, comprises vision transformer (ViT) architectures [4], a variant of transformer encoders [5], employing repeated transformer blocks divided into multi-head attention (MHA) and feedforward (FF) sub-blocks (Fig. 1). This architecture includes fully connected (FC) layers that capture long-range correlations across the entire image. State-of-the-art simulations indicate that both classes achieve comparable accuracies on small and medium-sized datasets, with ViTs potentially outperforming CNNs on large datasets by utilizing numerous learnable parameters that can exceed billions [6, 7].

Given the fundamental differences in their designs, the prevailing view is that the learning mechanisms underlying ViTs and CNNs are fundamentally distinct. Contrary to this intuition, recent work has demonstrated that both architecture types share a unified underlying learning mechanism [8]. This mechanism is based on quantifying single-nodal performance (SNP) using the matrix value of each node within each transformer encoder, analogous to single-filter performance (SFP) among CLs and SNP among FC layers in CNNs [9]. The results indicate that both CNN filters and ViT nodes identify small clusters of possible output labels, with additional noise represented by matrix elements outside these clusters. These features become increasingly refined along the layers of CNNs, enhancing the signal-to-noise ratio (SNR) and accuracy of the CLs. Additionally, SNP analysis has revealed the underlying learning mechanism of MHA in ViT architectures, which arises from emerging cooperation among nodes within each attention head along transformer blocks. This cooperation results in the emergence of spontaneous symmetry breaking among attention heads, where each head specializes in classifying a subset of labels. This phenomenon becomes increasingly pronounced in the final encoder blocks and is similar to phase transitions in statistical physics [10, 11].

The proposed symmetry breaking among the heads of MHA was demonstrated using the compact convolutional transformer architecture CCT-7/3x1, which comprised only four heads [12]. However, its applicability to large-scale MHA (LS-MHA) with an increased number of heads remains uncertain. Achieving High-accuracy image classification using CNNs or transformer architectures typically follows the relation in the last CL and transformer block:

$$No.\,of\,conv.filters\;O(10^3) > No.\,of\,labels > No.\,of\,attention\,heads\;O(1) \quad (1)$$

where each head consists of $\frac{MHA\,dim}{No.of\,heads} \cong O(10) - O(10^2)$ nodes, with $dim$ denoting the MHA dimension. Under these conditions, a subset of labels tends to dominate the accumulated SNP matrices associated with one head (e.g., appearing twice as often), leading to symmetry breaking among the heads [8]. This statistical mechanics inspired viewpoint enables characterization of the macroscopic behavior of each head and the entire network through the microscopic performance of each individual node's SNP. However, for a given MHA dimension and an increasing number of heads, LS-MHA, — comparable to the number of labels (e.g., 25 heads and 100 labels) — each head consists of a few nodes only:

$$\frac{LS-MHA\,dim}{No.of\,heads} = O(1) = \frac{LS-MHA\,dim}{No.of\,labels} \quad (2).$$

To generate a reliable signal at the output layer, each label must appear multiple times within clusters of the LS-MHA. However, following Eq. (2) it can be deduced that the possibility of each head specializing in a different subset of labels, thus producing symmetry breaking among many heads, is very unlikely. Such specialization might only be achievable in rare events by carefully adjusting the threshold of the matrix values. This study focuses on uncovering the underlying learning mechanism of LS-MHA based on a coarse-grained measure termed single-head performance (SHP), rather than SNP. This mechanism applies to both standard MHA and LS-MHA transformer architectures.

The results presented here are based on variants of the compact convolutional transformer CCT-7/3x1 [12], which uses a single CL as a tokenizer instead of image patching. This is followed by seven transformer encoder blocks and a classifier head composed of sequence pooling (SP) and a FC layer leading to the output layer (Fig. 2, left). The MHA dimension is $256 \times 256$, for a $32 \times 32$ input image size, followed by max pooling (Fig. 1). The number of heads in each transformer block is $H(i) = 4, i = 1, 2 \ldots, 7$.

This model achieved an accuracy of ~0.809 on the CIFAR100 dataset [13] after 1000 training epochs.

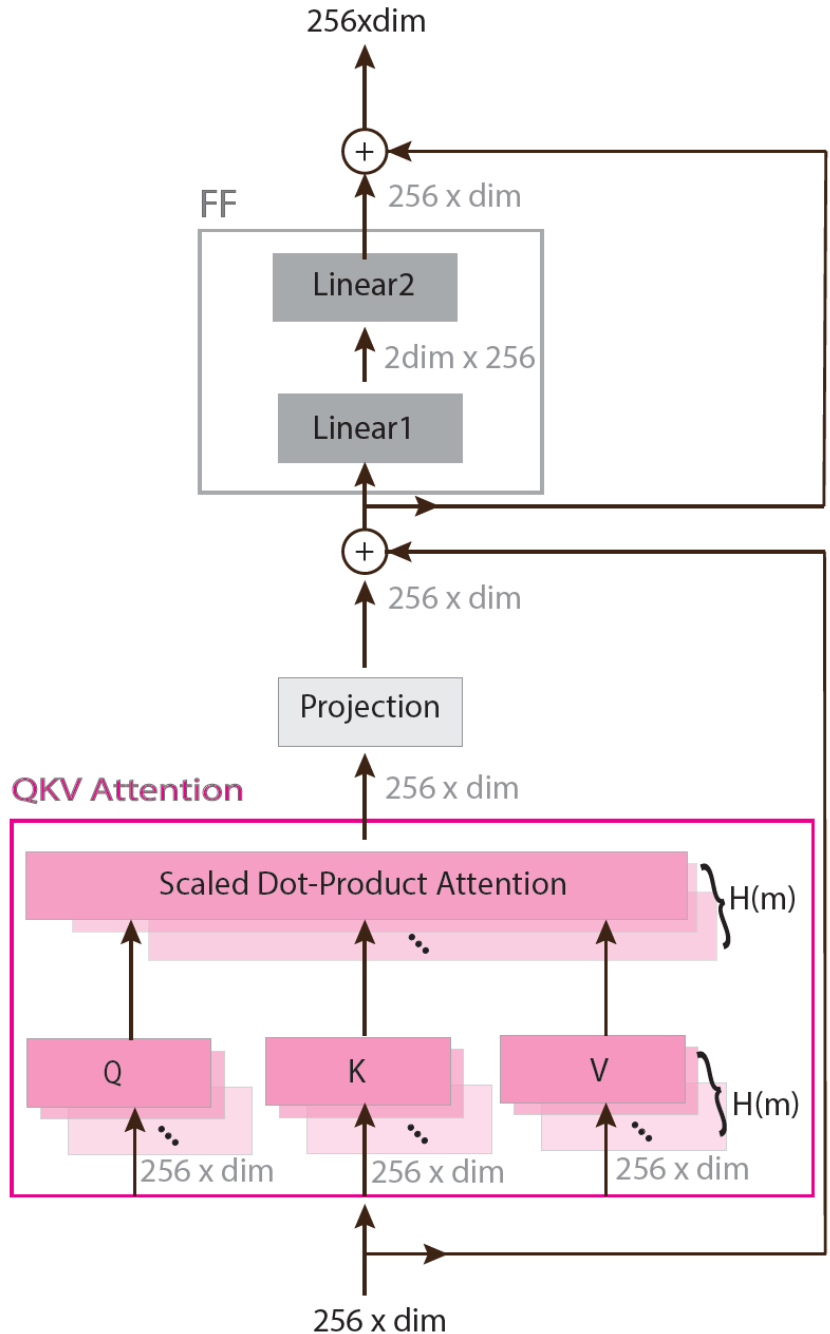

**Fig. 1.** Schematic of the $m^{th}$ generalized transformer encoder block in the CCT-7/3x1 architecture. It consists of QKV Attention with $H(m)$ heads (pink), followed by a FC projection layer (light gray), which together form the LS-MHA, where $dim$ denotes the

MHA dimension. This is followed by a feedforward (FF) sub-block composed of two FC layers (gray rectangles).

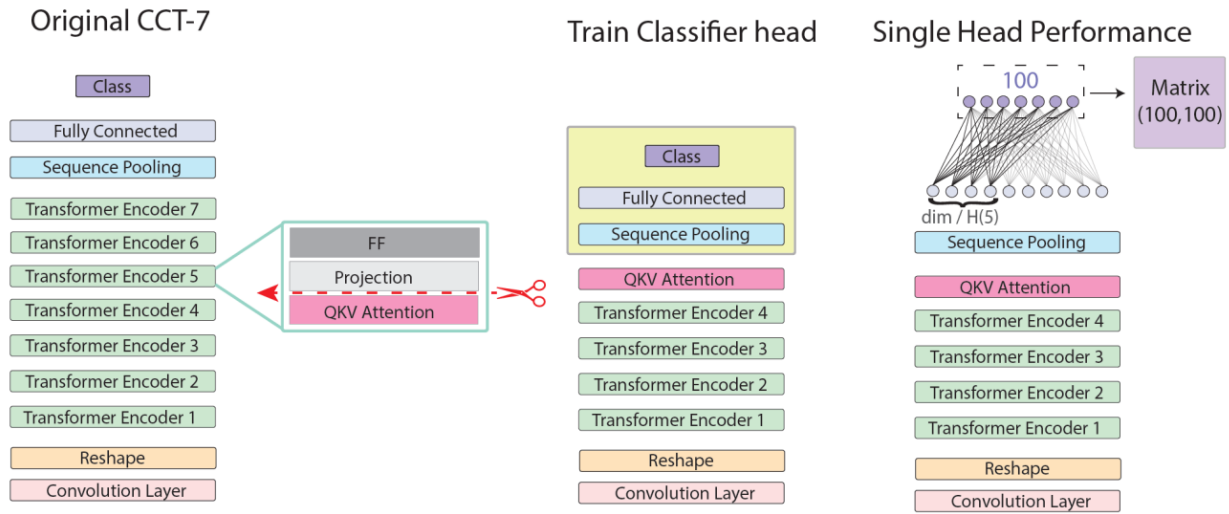

**Fig. 2.** Methodology for measuring single-head performance (SHP), demonstrated using the CCT-7/3x1 architecture. Left: Pre-trained CCT-7/3x1 on CIFAR-100, where each transformer block comprises three sub-blocks: QKV Attention (pink), Projection (light gray) and FF (gray) (see Fig. 1). The architecture is truncated after the QKV Attention of the $m^{th}$ transformer encoder (here exemplified as $m = 5$). Middle: Classifier head (yellow) is trained on CIFAR-100 to minimize loss function. Right: All weights in the FC layer are silenced (light gray) except those emerging from the selected head, consisting of $dim/H(5)$ nodes (black), where $dim$ is the LS-MHA dimension and $H(5)$ is the number of heads. The 100 output fields, averaged over each validation input label, generate a $100 \times 100$ matrix used to quantify the SHP.

## 2. Underlying ViT learning mechanism using SHP

The functionality of each of the seven transformer blocks in a pre-trained CCT-7/3x1 architecture [12] on the CIFAR-100 dataset [13] was quantified using two measurements, with the first $m$ transformer encoder blocks kept unchanged (frozen).

In the first measurement, the $256 \times \dim$ output units of the $m^{th}$ block—representing the partially preprocessed input image by the frozen portion of the CCT-7/3x1 architecture—were connected to the output layer using a classifier head composed of SP and a FC layer (Figs. 1 and 2). This classifier head was trained on CIFAR-100 to minimize the loss function, and the accuracy of this partial CCT-7/3x1 architecture was then estimated using the validation dataset, indicating a progressive increase with $m$ [8].

In the second measurement, the first $m - 1$ transformer blocks and the QKV Attention of the $m^{th}$ transformer encoder block in the pre-trained CCT-7/3x1 architecture remained frozen. The $256 \times dim$ output of its scaled dot-product attention was connected to a classifier head (Fig. 2, middle), trained in the same manner as in the first measurement. The accuracy of the LS-MHA was similarly estimated using the validation dataset, and also showed a progressive increase with $m$ (Table 1).

The functionality of each of the $H(m)$ heads was estimated by silencing all input nodes to the trained classifier head except those belonging to the examined head (Fig. 2). Consequently, the 100 output units representing the labels were influenced only by $dim/H(m)$ nodes of that head. The validation dataset was then propagated through the first $m - 1$ pre-trained transformer blocks, as well as through the silenced QKV Attention of the $m^{th}$ block, generating a $100 \times 100$ value matrix. Each matrix element $(i, j)$ represents the average field generated on output unit $j$ by validation inputs with label $i$, normalized by the maximum matrix element (Fig. 3, left). A Boolean clipped matrix was then derived by applying a threshold (Fig. 3, middle), followed by a permutation to form diagonal clusters (Fig. 3, right). Elements above the threshold but located outside these diagonal clusters were classified as noise $n$ (yellow).

The matrix representing the first SHP among the four heads $H(7) = 4$ of the 7[th] transformer block was exemplified for thresholds of $0.3$ and $0.6$ (Fig. 4). These matrices are almost identical to the matrices obtained by averaging the $64$ SNP matrices of the same head, followed by clipping with the same threshold (Fig. 4). Moreover, at the threshold 0.6, both methods, the SHP and head performance (HP) derived from SNP, exhibited identical modus vivendi across the heads, with each head recognizing nearly the same subset of labels (Fig. 4) and exhibiting negligible noise, $n$ (Table 1). However, the similarity between the symmetry breaking observed using SHP and HP via SNP depends critically on careful tuning of the threshold according to the properties of the architecture and the chosen definition of symmetry breaking. The average properties of the four SHP matrices across transformer blocks [1,7] are summarized in Table 1, revealing consistent trends: the accuracy of the QKV Attention module, $Attn.Acc.$, and the SNR progressively increasing along the transformer blocks, paralleling the accuracy trend measured at the output of each transformer block in the first measurement.

| Block | $Attn.Acc.$ | $N_c$ | $C_s$ | Diag | $n$ | $N_{label}$ | $N_{noise}$ | $N_{inter}$ | SNR |
|---|---|---|---|---|---|---|---|---|---|
| 7 | 0.780 | 36.5 | 1.03 | 37.5 | 12.5 | 1.5 | 0.005 | 0.0004 | 277.2 |
| 6 | 0.728 | 45.2 | 1.07 | 48.5 | 39.7 | 1.9 | 0.016 | 0.0014 | 112.2 |
| 5 | 0.676 | 55.0 | 1.11 | 61 | 53 | 2.4 | 0.021 | 0.0027 | 102.2 |
| 4 | 0.635 | 59.7 | 1.21 | 72.5 | 146 | 2.9 | 0.058 | 0.0062 | 44.9 |
| 3 | 0.578 | 57.2 | 1.23 | 70.5 | 198.5 | 2.8 | 0.079 | 0.0065 | 32.8 |
| 2 | 0.442 | 29.2 | 1.29 | 38.0 | 220.5 | 1.5 | 0.088 | 0.0045 | 16.4 |
| 1 | 0.309 | 25.7 | 1.39 | 36.0 | 235.2 | 1.4 | 0.094 | 0.0057 | 14.4 |

**Table 1.** Statistical properties of SHP matrices using a threshold of 0.3, based on a classifier head trained on the QKV Attention modules of the transformer blocks within the pre-trained CCT-7/3x1 model on the CIFAR-100 datasets. Each block comprises $H(i) = 4$ attention heads, achieving an overall accuracy of approximately ~0.809. The table reports the accuracy of the QKV Attention, $Attn.Acc.$, average number of clusters per SHP matrix, $N_c$, average cluster size, $C_s$, average number of diagonal elements per SHP

matrix, $Diag$, average noise per SHP matrix, $n$, average number of label appearances across all heads of a MHA, $N_{label}$, average noise per matrix element averaged over all SHP matrices, $N_{noise}$, and average internal noise per label averaged over all SHP matrices, $N_{inter}$.

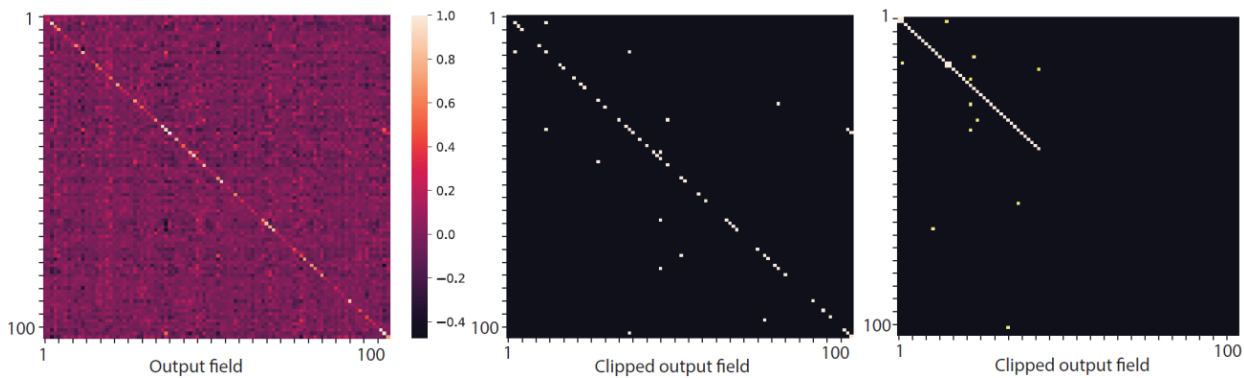

**Fig. 3.** Exemplified SHP for CCT-7/3x1. Left: Matrix value generated by $dim/H(7)$ nodes ($H(7) = 4$) of the seventh QKV Attention block on the 100 output units using a trained classifier head (Fig. 2, middle), with all other heads silenced (Fig. 2, right). Element $(i, j)$ represents the averaged fields generated by label $i$ validation inputs on an output unit $j$, with matrix elements normalized by their maximum value. Middle: Boolean clipped matrix using a threshold of 0.3. Right: Permuted clipped matrix showing ~40 block diagonal clusters of size 1, except for one cluster of size 2, and a few above-threshold elements outside clusters (yellow).

At threshold 0.3, the cluster size, $C_s$, remains close to unity regardless of transformer block number, while the number of clusters per head decreases from about half the labels (50) to 1/3 (33) along the transformer blocks. This reflects a low label signal, as each label appears on average only $N_{lable} = 1.5$ times across the four SHP matrices (Table 1). Consequently, internal cluster noise, $N_{inter}$, the average number of appearances of other labels within the clusters forming the signal of a given label, is negligible owing to $C_s \sim 1$. To achieve high $SNR$, the external noise, $n$, must be significantly lower than the signal.

Assuming external noise is homogeneously distributed over ~10,000 out-of-cluster elements, the $SNR = \frac{signal}{N_{noise}+N_{inter}} \gg 1$ increases along the transformer blocks (Table 1). This homogeneous noise assumption may be less informative when $\frac{noise}{No.\ matrix\ elements} \ll 1$, nevertheless, the increased $SNR$ along the transformer blocks supports the learning mechanism.

For the same CCT-7/3x1 architecture with four heads in the first six blocks, $H(i) = 4\ i = 1, ..., 6$, and varying heads in the seventh block, $H(7) = 4,\ 8,\ 16,\ \text{or}\ 32$, the accuracy remains ~0.81 independent of $H(7)$. External noise, $N_{noise}$, typically increases with $H(7)$ because each head contains fewer nodes (Table 2). However, the overall LS-MHA accuracy $Attn.Acc.$, remains stable, as the increased noise is offset by the higher number of heads (Table 2).

| H(7) | $H(7)_{size}$ | Acc. | Attn. Acc. | $N_C$ | $C_S$ | Diag | n | $N_{label}$ | $N_{noise}$ | $N_{inter}$ | SNR |
|---|---|---|---|---|---|---|---|---|---|---|---|
| 4 | 64 | 0.809 | 0.780 | 36.5 | 1.03 | 37.5 | 12.5 | 1.5 | 0.0050 | 0.0004 | 277.2 |
| 8 | 32 | 0.806 | 0.792 | 30.5 | 1.06 | 32.4 | 11.1 | 2.6 | 0.0089 | 0.0016 | 246.8 |
| 16 | 16 | 0.810 | 0.777 | 15.4 | 1.17 | 18.0 | 55.3 | 2.9 | 0.0885 | 0.0051 | 30.9 |
| 32 | 8 | 0.807 | 0.772 | 11.8 | 1.30 | 15.4 | 139.3 | 4.9 | 0.4458 | 0.0149 | 10.7 |

**Table 2.** Accuracy and statistical properties of CCT-7/3x1 architectures with $dim = 256$, four heads in first six $blocks, H(i) = 4\ i = 1, ..., 6$, and varying numbers of heads in the last block $H(7) = 4,\ 8, 16,\ \text{and}\ 32$, where $H(7)_{size}$ denotes the size of each head. Notations are consistent with those in Table 1.

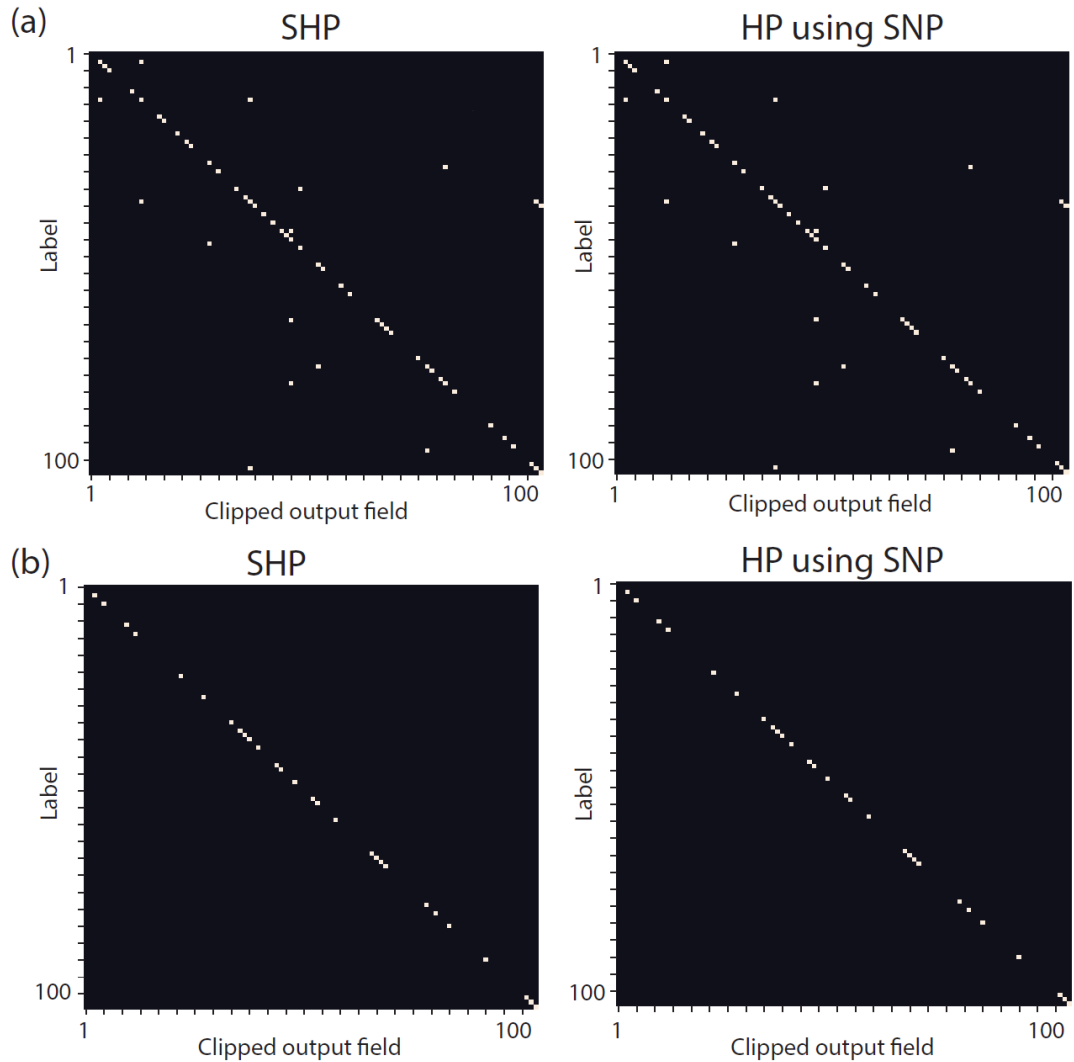

**Fig. 4**. Comparison of SHP and HP obtained by averaging over the SNP matrices belonging to the same head. (a) SHP (left) and HP using 64 SNP matrices (right) for the first head of $H(7)$, in CCT-7/3x1 trained on CIFAR-100 (with four heads for all MHA and $dim = 256$), using threshold 0.3. (b) Same as panel (a), but using threshold 0.6.

The sensitivity of the accuracy and its maximum value as a function of the number of heads was investigated in CCT-1/3x1, comprising a single CL as a tokenizer instead of image patching, followed by a single transformer encoder block with $dim = 256$ connected to the output layer using a classifier head. Training this architecture on

CIFAR-100 with varying numbers of heads indicates that maximum accuracy is achieved at $H(1) \sim 16$, where the head size is also 16 (Table 3). For the same architecture but with LS-MHA $dim = 1024$, the maximum accuracy is higher and achieved at $H(1) = 64$, with head size remaining 16 (Table 3). In these shallow architectures dominated by one CL and one transformer, increasing $dim$ from 256 to 1024 (Table 3) improves accuracy by approximately 5%, and is attributed to the increased number of filters in the CL, as previously observed for CNNs [14-16]. However, in deeper CCT architectures, primarily composed of transformer blocks, accuracy sensitivity to $dim$ is diminished (Tables 4–5).

The accuracy of CCT-1/3x1 (~0.71) is approximately 10% lower than that of CCT-7/3x1 (~0.81). However, the SHP properties of CCT-1/3x1 indicate that the cluster size $C_s$ approaches unity and the noise levels are similar to those of the initial transformer blocks of CCT-7/3x1 (Table 1).

| $H$ | 1 | 2 | 4 | 8 | 16 | 32 | 64 |
|---|---|---|---|---|---|---|---|
| $H_{size}$ | 256 | 128 | 64 | 32 | 16 | 8 | 4 |
| Acc. | 0.647 | 0.658 | 0.777 | 0.692 | 0.698 | 0.690 | 0.677 |

| $H$ | 1 | 2 | 4 | 8 | 16 | 32 | 64 | 128 |
|---|---|---|---|---|---|---|---|---|
| $H_{size}$ | 1024 | 512 | 256 | 128 | 64 | 32 | 16 | 8 |
| Acc. 300 | 0.611 | 0.648 | 0.672 | 0.689 | 0.712 | 0.719 | 0.725 | 0.720 |
| Acc. 1000 | 0.641 | 0.698 | 0.711 | 0.733 | 0.738 | 0.746 | 0.751 | 0.745 |

**Table 3.** Accuracy of CCT-1/3x1 trained on CIFAR-100 with varying number of heads, $H$, and size $H_{size}$, such that $H \cdot H_{size} = dim$. Upper Table: $dim = 256$ using 1000 epochs. Lower Table: $dim = 1024$ with accuracy achieved using 300 (first row) and 1000 epochs (second row).

The underlying learning mechanism of an LS-MHA transformer architecture does not rely on symmetry breaking among the heads, unlike its counterpart architecture with MHA [8]. Analysis using the SHP metric indicates that each head explicitly recognizes a finite subset of labels with unit cluster size and negligible noise. These subsets of labels recognized by each one of the heads are partially overlap, causing each label to appear in a minority of all SHP matrices, thus forming a weak signal but significantly stronger than the average noise per matrix element, $N_{noise}$. This generalized underlying mechanism also applies to architectures with a small number of heads, such as MHA. In addition, it quantitatively differs from the learning mechanism of CNNs, where the number of filters is much larger than the number of heads and typically exceeds the number labels. In CNNs, each label appears frequently across clusters of SFP matrices in a CL, but noise and internal cluster noise are also enhanced. Nevertheless, CNNs lack a natural grouping of filters into meaningful entities analogous to the SHP, which is composed of SNPs belonging to a head in transformer architectures.

### 3. Advanced accuracy with low latency

The result that the average cluster size, $C_s$, approaches unity for SHP across all seven transformer blocks of CCT-7/3x1 (Table 1) and even for CCT-1/3x1 differs with the behavior observed in CNN [9, 17, 18]. This suggests that replacing the first several transformer blocks with CLs may not adversely affect accuracy, as $C_s$ is already near unity, while noise is expected to decrease along the initial CLs.

   An example of such an architecture is CCT-2/3x5, which comprises five initial CLs (including padding to preserve spatial dimensions and max pooling after only the first CL), followed by two transformer blocks. The total number of layers in this architecture, which correlates directly with network latency, is $5 + 4 \times 2 + 1 = 14$, substantially fewer than $1 + 7 \times 4 + 1 = 30$ layers in CCT-7/3x1. The achieved accuracy for CCT-2/3x5, with the number of heads varying between $4$ and $32$ and over $1000$ training epochs, is ~0.81 and ~0.82 for $dim$ 256 and 512, respectively, comparable to CCT-7/3x1 (Table 4). These results clearly indicate that similar accuracy can be obtained with architectures of the same type but with significantly reduced latency and fewer tunable parameters. Moreover, increasing the $dim$ by a factor of four yields only a mild improvement in accuracy (Table 4) and shows near saturation beyond 1024 (not shown). This saturation contrasts with CNNs, where proportionally increasing the number of filters in CLs reduces error rates following a power law, indicating wide-shallow learning as an alternative to deep learning [14-16].

It is also noteworthy that reversing the order of CLs and transformer blocks markedly reduces accuracy. For instance, a CCT architecture with an initial CL tokenizer connected to two transformer blocks followed by four CLs achieves accuracy well below 0.8, despite consisting of the same components as CCT-2/3x5. This demonstrates that learning short-range correlations directly from input data is significantly more efficient than performing this operation after spatial mixing via FC layers of transformers.

| $H$ | $H_{size}$ | $Dim$ | Acc. 300 | Acc. 1000 |
|---|---|---|---|---|
| 4 | 64 | 256 | 76.07 | 80.70 |
| 16 | 16 | 256 | 77.26 | 81.04 |
| 32 | 16 | 512 | 78.81 | 82.18 |
| 64 | 16 | 1024 | 79.60 | 82.12 |

**Table 4.** Accuracy of CCT-2/3x5 architectures trained for $300$ and $1000$ epochs with the number of heads $H(1) = H(2) = H$ varying in the range $[4, 64]$, and $dim$ in the range $[256, 1024]$.

### 4. Soft committee decision and enhanced accuracy

Results revealed a notable difference between classification using ViT architectures (Tables 2–4) and CNNs. Several CCT architectures comprising of seven components, with varying numbers of CLs, transformer blocks, heads, and $dim$, achieve similar accuracies of ~0.81 (Tables 2–4). This phenomenon is largely absent in CNNs, where altering the number of CLs or filters per CL typically causes significant changes in the accuracy [14-16].

Different ViT architectures with comparable accuracies differ in the properties of their SHP matrices, suggesting potential accuracy improvements through a soft committee decision approach. Each validation input processed by a trained CCT architecture on CIFAR-100 produces a 100-dimensional output vector; for a soft committee of $N$ CCT architectures, the final decision is based on the sum of these output vectors. It is similar to soft voting and bagging approaches [19, 20], however, performance gain stems primarily from architectural diversity resulting in different functionality of the heads.

For a committee of only $N = 4$ CCT architectures—each composed of seven components (CLs and transformer blocks) and achieving an average independent accuracy of 0.813 (Table 5)—the soft committee raises accuracy to 0.851. This enhancement of approximately 4% for $N = 4$ is remarkable, particularly when compared to a soft committee of $N = 4$ CCT-7/3x1 trained from random initial conditions, which improved accuracy from ~0.809 to ~0.822 only. The similarity between the $N = 4$ trained architectures is measured by their average agreement on the validation set, defined as the probability that a pair of architectures simultaneously predict correctly or incorrectly. For uncorrelated architectures with accuracy 0.81, this probability is expected to be $0.81^2 + 0.19^2 = 0.692$, while for fully correlated outputs it equals 1. The observed average agreement among the four architectures (Table 5) is ~0.87, whereas for the CCT-7/3x1 architectures trained from random initial conditions it is higher, ~0.885, supporting the difference in accuracy gains between the two committee scenarios. These findings suggest that asymptotically, $N \gg 1$, soft committee accuracy can be further enhanced by combining diverse architectures especially with low correlation.

| CL | No. Trans | $H$ | $H_{size}$ | Dim | Acc. | Acc. head | $N_c$ | $C_S$ | Diag | $n$ | $N_{label}$ | $N_{nois}$ | $N_{inter}$ | SNR |
|---|---|---|---|---|---|---|---|---|---|---|---|---|---|---|
| 1 | 7 | 4 | 64 | 256 | 0.809 | 0.780 | 36.5 | 1.02 | 37.5 | 12.5 | 1.5 | 0.1 | 0.0004 | 14.8 |
| 1 | 7 | 4 H(7)=16 | 64 H(7)=16 | 256 | 0.813 | 0.753 | 16.5 | 1.07 | 17.7 | 74.1 | 2.8 | 0.1 | 0.0022 | 23.5 |

| 5 | 2 | 16 | 16 | 256 | 0.810 | 0.752 | 16.5 | 1.13 | 18.6 | 65.9 | 2.9 | 0.1 | 0.0038 | 27.3 |
| 5 | 2 | 32 | 16 | 512 | 0.822 | 0.787 | 14.7 | 1.25 | 18.4 | 122.0 | 5.9 | 1.0 | 0.0149 | 5.8 |

**Table 5.** Four CCT architectures with an average independent accuracy of $0.813$ and a soft committee of $\sim 0.851$. Each architecture is characterized by the number of CLs followed by the number of transformer blocks, $No.Trans$, with each transformer block comprising $H$ heads, except for the second architecture (second row) where the seventh block consists of $16$ heads. Other notations follow those in Table 1.

Latency can be further reduced, while maintaining the accuracy of CCT-7/3x1, by employing a soft committee of CCT-2/3x2 architectures, each composed of two CLs, followed by two transformer blocks. The latency of these architectures is $2 + 4 \times 2 + 1 = 11$, compared to a latency of $30$ for CCT-7/3x1. Using LS-MHA with $dim$ $512$ and $1024$ and varying the number of heads between $8$ and $64$ for each one of the two transformer blocks, the average independent accuracy of six CCT-2/3x2 architectures is $\sim 0.79$ (Table 6). A soft committee of $N = 5$ and $6$ yields accuracies of $\sim 0.827$ and $0.83$, respectively (Table 6), surpassing even the reported accuracy for CCT-7/3x1 trained over $5000$ epochs[12]. These results demonstrate that a soft committee comprising different shallow ViT architectures can imitate the accuracy of a much deeper single ViT architecture. However, this soft committee mechanism differs fundamentally from the efficient shallow-learning mechanism in CNNs, which relies on increasing the number of filters in the CLs [14-16], thus creating wide-shallow CNN architectures.

The latency can be further reduced, while approximating the accuracy of CCT-7/3x1, by employing a soft committee of architectures, each containing a single transformer block. These architectures consist either of one CL followed by one transformer block, CCT-1/3x1, or two CLs followed by one transformer block, CCT-1/3x2. The total latency for CCT-1/3x1 is $1 + 4 + 1 = 6$, while for CCT-1/3x2 it is $2 + 4 + 1 = 7$, both of which are less than one quarter of the latency of CCT-7/3x1, which is $30$. A soft committee of $N =$

10, combining five CCT-1/3x2 and five CCT-1/3x1 architectures, results in ~0.811 accuracy (Table 7), comparable to the 0.809 accuracy obtained by CCT-7/3x1 after 1000 epochs. Although the average independent accuracy of CCT-1/3x2 exceeds that of CCT-1/3x1 by 0.017 (Table 7), the soft committee of $N = 10$ CCT-1/3x2 architectures alone achieves only 0.802 accuracy. This improvement in soft committee accuracy for the combined CCT-1/3x1 and CCT-1/3x2 architectures is attributed to their lower correlation on the validation set. Specifically, the correlation between pairs of CCT-1/3x1 or pairs of CCT-1/3x2 architectures is on average ~0.02 higher than the correlation between CCT-1/3x1 and CCT-1/3x2 architectures. Thus, maximizing soft committee accuracy for a fixed $N$ does not equate to maximizing the independent accuracy of each constituent architecture. These results suggest that a soft committee composed of architectures with varying latencies can further enhance accuracy, beyond combinations of identical architectures with different LS-MHA configurations. A comparison of results from Tables 6 and 7 also reveals a trade-off between latency and committee size $N$, indicating that reducing latency generally requires increasing $N$ to maintain a given accuracy.

| $H$ | $H_{size}$ | $Dim$ | Acc. | Acc. head |
|---|---|---|---|---|
| 64 | 16 | 1024 | 0.799 | 0.764 |
| 32 | 32 | 1024 | 0.805 | 0.765 |
| 64 | 8 | 512 | 0.788 | 0.747 |
| 8 | 64 | 512 | 0.786 | 0.721 |
| 16 | 32 | 512 | 0.792 | 0.749 |
| 32 | 16 | 512 | 0.793 | 0.751 |

**Table 6.** Six CCT-2/3x2 architectures with an average independent accuracy of ~0.79 and a soft committee accuracy of ~0.83. The soft committee accuracy excluding the last architecture (first five) is 0.827.

| No. CLs | 1 | 1 | 1 | 1 | 1 | 2 | 2 | 2 | 2 | 2 |
|---|---|---|---|---|---|---|---|---|---|---|
| $H$ | 64 | 64 | 192 | 93 | 128 | 64 | 64 | 192 | 93 | 78 |
| $H_{size}$ | 24 | 16 | 16 | 11 | 16 | 24 | 16 | 16 | 11 | 13 |
| $dim$ | 1536 | 1024 | 3072 | 1023 | 2048 | 1536 | 1024 | 3072 | 1023 | 1014 |
| No. epochs | 1000 | 3000 | 1000 | 1000 | 1000 | 1000 | 2000 | 1000 | 2000 | 2000 |
| Acc. | 0.757 | 0.756 | 0.755 | 0.754 | 0.754 | 0.773 | 0.772 | 0.775 | 0.769 | 0.774 |

**Table 7.** Five CCT-1/3x1 architectures with an average independent accuracy of ~0.752, and five CCT-1/3x2 architectures with an average independent accuracy of ~0.773. The soft committee of these $N = 10$ systems achieves an accuracy of ~0.811.

## 5. Discussion

All types of machine learning models comprise nodes and weights; however, their fundamental computational entities differ: filters for CNNs and heads for transformer architectures implementing LS-MHA. During training, each of these entities develops correlations among its internal weights and nodal functionality, as measured by the underlying learning mechanism. Specifically, this mechanism reveals that each fundamental entity tends to identify a small cluster of output labels associated with a given input label. An entity may comprise several such clusters, each representing a limited subset of possible output labels. The number of times a label appears across all entities within a CL or LS-MHA module constitutes a signal. In this framework, two forms of noise are defined. External noise represents above-threshold appearances of labels outside the identified clusters, while internal cluster noise arises from the inclusion of non-target labels within the clusters themselves. These quantities collectively determine the SNR, which typically increases with the layers of the architecture.

Although this learning mechanism applies broadly, it exhibits quantitative differences between CNNs and transformer-based architectures, reflecting the distinct properties of their computational entities (see Eq. (1)). In CNNs—particularly in the final CLs—the number of filters is usually on the order of $O(10^3)$, typically exceeding the number of classification labels. In contrast, the number of heads in a standard MHA is typically $O(1)$, and in LS-MHA, it is generally $O(10)$. In transformer architectures, each attention head tends to form clusters of size approaching unity, and the number of such clusters is typically a significant fraction of the total number of labels. These clusters exhibit vanishing internal noise and low external noise, both of which decrease along successive blocks. This allows a relatively small number of heads to generate distinct signals for each label, thereby achieving a high SNR. In contrary, filters in CNNs often form a smaller number of clusters, which may be relatively large—sometimes exceeding 15 labels per cluster [18]. This leads to greater internal noise, and external noise may be an order of magnitude higher than that observed in transformer heads [9, 17, 18]. Consequently, the comparatively limited information captured by each filter is offset by deploying a much larger number of filters than attention heads in transformer architectures.

Increasing the number of filters per layer in CNNs results in a power-law decay of error rates toward zero, indicating an efficient wide-shallow learning mechanism as an alternative to deep learning [15, 16]. A comparable decay was not observed in the examined transformer architectures, whether by increasing the number of heads for a given $dim$ or by increasing $dim$ while keeping the number of heads constant (Tables 2–7). One possible conclusion is that wide-shallow transformer architectures may not replicate the accuracy achieved by deep learning. However, it remains possible that such a decay occurs at much larger values of $dim$, beyond the regime influenced by the double-gradient phenomenon [21], or when both $dim$ and the number of heads increase simultaneously and vary across transformer blocks. Investigating these directions requires optimizing complex tasks with large transformer blocks, which exceeds our current computational capabilities.

Replacing each of the initial several transformer blocks in the examined architecture with CLs can significantly reduce latency and the number of trainable parameters without compromising accuracy (Table 4). This suggests the potential of an efficient hybrid architecture wherein the initial phase of learning — "ignition" — is realized by a CNN, while the subsequent accuracy elevation is achieved by transformer block encoders. The optimal ratio between these two consecutive components of hybrid deep learning—ignition and accuracy elevation—is yet to be discovered. Nevertheless, the results indicate that such hybrid learning is not commutative, as a significant decrease in accuracy occurs when ignition is performed by transformer blocks followed by accuracy enhancement via CNNs. Furthermore, convolutions are less effective when the input data is mixed by FC layers of the transformers.

The possibility of achieving comparable accuracy using hybrid architectures and architectures with varying $dim$ and number of heads enables a significant accuracy improvement through soft committees composed of such architectures. To further enhance accuracy, a larger ensemble of different architectures — each achieving similar accuracy but exhibiting minimal average agreement on the validation set — is required. This objective is not equivalent with maximizing the independent accuracy of each individual architecture. The results indicate that soft committees comprising architectures with differing latencies can enhance accuracy beyond that obtained from combinations of the same architecture configured with different LS-MHA structures. These findings suggest that the performance of deep transformer architectures can be effectively approximated by a soft committee of substantially shallower transformer architectures. A deeper understanding of the underlying learning mechanism across different ViT architectures [22-25] may guide the construction of such architecture ensembles and the determination of the asymptotic accuracy achievable by soft committees with many members. This form of shallow-learning mechanism differs fundamentally from that of CNNs, which rely on increasing the number of filters within a fixed architecture. The possibility of efficient training of multiple transformer architectures as a single soft committee decision, sharing a common output layer, represents a promising direction. This approach may further enhance accuracy by splitting learning of each input by partial committee members, analogous to multilayer perceptrons [26].

The results and conclusions presented herein are based on a limited subset of ViT architectures and warrant generalization and validation across other transformer classes as well as other datasets such as ImageNet consisting of larger images [27]. The quantitative differences between the learning mechanisms of CNNs and transformers — especially the prevalence of unit-sized clusters per head with negligible internal noise — are expected to play a crucial role in natural language processing applications [5]. The optimal number and dimensions of heads in the LS-MHA are expected to scale with the number of labels and classification decisions [28-31]. Moreover, the development of tailored backpropagation procedures for individual heads to direct cluster formation and thereby enhance accuracy presents an intriguing avenue for future research.

Finally, it is possible that the proposed underlying learning mechanism approach might impact the analysis of other supervised and unsupervised learning based on shallow architectures [10, 32-34], as well as adaptive networks of dynamical systems such as network physiology [35-37].

**Acknowledgements**

The work is supported by the Israel Science Foundation [grant number 346/22].

**Appendix**

*Dataset and preprocessing:* The datasets used in this study is CIFAR-100 [13]. Each pixel value was normalized by subtracting the mean and dividing by the standard deviation of its image. In all simulations, data augmentation derived from the original images was performed, by using CutMix[38], Mixup[39], Randaugment[40], and Random Erasing[41].

*Optimization:* The hyper-parameters $\eta$ (learning rate) and $\alpha$ (L2 regularization[42]) were optimized for offline learning, using a mini-batch size of 128 inputs. Both the learning rate and the weight decay were reduced per epoch based on cosine annealing[43] until a linear scheduler is mentioned such that $\eta$ was multiplied by the decay factor, $q$, every $\Delta t$ epochs, and is denoted as $(q, \Delta t)$. The Soft Target Cross-Entropy function[44] was selected for the classification task and minimized using the stochastic gradient decent algorithm [3, 45]. The AdamW optimizer was used[46]. The minimal loss was determined by searching through the hyper-parameters (see below).

*Hyper- parameters:* The hyper-parameters used for Table 1, Table 2 and Fig. 3. are $\eta = 6e-4, \alpha = 6e-2$ and the architecture was trained for 1000 epochs. The hyper-parameters used for the classifier head are $\eta = 1e-3, \alpha = 6e-2$, a linear scheduler $(0.78, 10)$ was applied, and was trained for 100 epochs. The hyper-parameters used for Table 3 are $\eta = 6e-4, \alpha = 6e-2$ and the architecture was trained 1000 epochs for the upper table and 300 and 1000 epoch for the lower table. The hyper-parameters used for the classifier head are $\eta = 1e-3, \alpha = 6e-2$, a linear scheduler $(0.78, 10)$ was applied, and was trained for 100 epochs. The hyper-parameters used for Table 4 are $\eta = 6e-4, \alpha = 6e-2$ and the architecture was trained 300 and 1000 epochs. The hyper-parameters used for Tables 5-7 are $\eta = 6e-4, \alpha = 6e-2$ and the architecture was trained for 1000 epochs, unless stated otherwise. The hyper-parameters used for the classifier head are $\eta = 1e-3, \alpha = 6e-2$, a linear scheduler $(0.78, 10)$ was applied, and was trained for 100 epochs.

*Soft Committee:* The soft committee decision was performed by the summation of all output fields without any alterations such as Softmax, activation or normalization. The decision was then made on the summed field.

*Statistics:* Statistics for all results were obtained using at least three samples and the standard division was less than $0.5\%$ for all the results.

*Hardware and software:* We used Google Colab Pro and its available GPUs. We used Pytorch for all the programming processes.